\documentclass[letterpaper, 10 pt, conference]{ieeeconf}  %

\IEEEoverridecommandlockouts                              %

\usepackage{amsmath}
\usepackage{tikz}
\usetikzlibrary{shapes,arrows, calc, arrows.meta, intersections, positioning, patterns, decorations.pathreplacing, backgrounds} %
\usepackage{selectp} %
\usepackage{mathtools}
\usepackage{subcaption}
\usepackage{siunitx}
\usepackage{cite}
\usepackage{makecell, multirow}
\usepackage{hyperref}
\setcellgapes{1.5pt}
\usepackage{booktabs}
\usepackage[absolute,overlay]{textpos}

\newcommand*\rot{\rotatebox{90}}
\definecolor{TABLE_GOOD}{RGB}{0,100,0}
\definecolor{TABLE_BAD}{RGB}{214,39,40}
\def\linkToCode{www.github.com/translearn/limits}

\title{\LARGE \bf
Learning Robot Trajectories subject to Kinematic Joint Constraints
}

\author{Jonas C. Kiemel$^{1}$ and Torsten Kröger%
\thanks{$^{1}$Institute for Anthropomatics and Robotics – Intelligent Process Automation and Robotics (IAR-IPR),
	Karlsruhe Institute of Technology (KIT),
	jonas.kiemel@kit.edu, $^{2}$\url{\linkToCode}
	}%
}

\begin{document}

\maketitle
\thispagestyle{empty}
\pagestyle{empty}

\begin{textblock*}{14.9cm}(3.2cm,0.75cm) 
	{\footnotesize © 2021 IEEE.  Personal use of this material is permitted.  Permission from IEEE must be obtained for all other uses, in any current or future media, including reprinting/republishing this material for advertising or promotional purposes, creating new collective works, for resale or redistribution to servers or lists, or reuse of any copyrighted component of this work in other works.}
\end{textblock*}

\begin{abstract}

We present an approach to learn fast and dynamic robot motions without exceeding limits on the \mbox{position $\theta$}, \mbox{velocity $\dot{\theta}$}, acceleration $\ddot{\theta}$ and jerk $\dddot{\theta}$ of each robot joint.
Movements are generated by mapping the predictions of a neural network to safely executable joint accelerations.
The neural network is invoked periodically and trained via reinforcement learning.
Our main contribution is an analytical procedure for calculating safe joint accelerations, which considers the prediction frequency $f_N$ of the neural network. 
As a result, the frequency $f_N$ can be freely chosen and treated as a hyperparameter.
We show that our approach is preferable to penalizing constraint violations as it provides explicit guarantees and does not distort the desired optimization target. 
In addition, the influence of the selected prediction frequency on the learning performance and on the computing effort is highlighted by various experiments. 

\end{abstract}

\section{INTRODUCTION}

 Over the last few decades, robots have become increasingly prevalent in the manufacturing industry.
   While industrial robots are superior to humans in terms of speed and precision, they lack the capability to react to unforeseen events. For instance, human workers can easily deal with elastic or varying workpieces, whereas robots rely on precise models of their environment. 
    Tracing back to remarkable achievements in gaming domains \cite{mnih2013playing, silver2017mastering}, model-free reinforcement learning (RL) has caught the attention of researchers trying to address the problem of flexible industrial production \cite{levine2018learning, kalashnikov2018scalable, berscheid2020, thomas2018learning, luo2019reinforcement, kiemel2020trueRMA, inoue2017deep,  schoettler2019deep, kaspar2020sim2real}. %
    Most commonly, the learning process involves a neural network that receives sensor data as input and outputs actions to parameterize motions. 
    In simulated environments, various action parameterizations (e.g. target joint positions, velocities or torques) have proven to perform well \cite{peng2017learning}. 
    However, when learning trajectories for real robots, the predicted motions must be executable without overloading the robot joints.
    In particular, jerk constraints are often ignored, although unbounded jerks are known to decrease the  life span of the robot joints and to stimulate natural frequencies \cite{Ibarz_2021, kroger2010line}. 
    
\begin{figure}[t]
	\includegraphics[trim=0 8 0 0, clip, width=\linewidth]{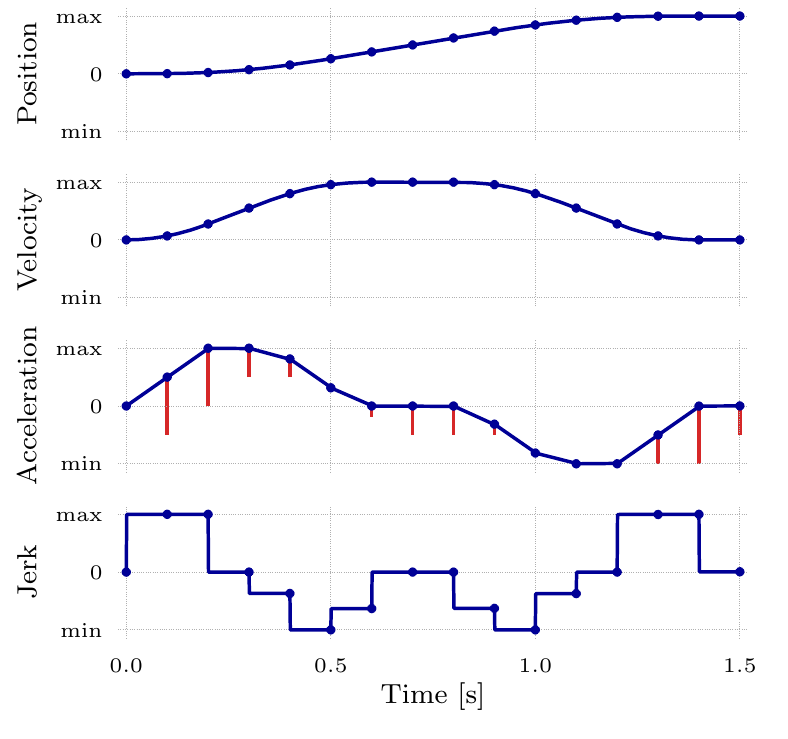}
	\caption{
	Our method enables learning of robot trajectories that do not violate position, velocity, acceleration and jerk limits. 
	 At each decision step, a neural network predicts an \mbox{action $\in [-1, 1]$} that is linearly mapped along the range of safe accelerations (red line).
	Prediction frequency  $f_N$: \SI{10}{\hertz}}
	\label{fig:max_trajectory_example}
	\vspace{-0.4cm}
\end{figure}
    Fig. \ref{fig:max_trajectory_example} illustrates our approach to limit the position, velocity, acceleration and jerk based on an exemplary trajectory for a single joint. 
    At each discrete decision step, the desired acceleration at the beginning of the following interval is determined. 
    As a first step, we calculate the range of accelerations that can be safely executed. 
    In Fig. \ref{fig:max_trajectory_example}, this range is visualized by a red line.
    Secondly, a neural network predicts an action $\in [{-1}, 1]$ that is linearly mapped along the range of feasible accelerations. In our example, every action is \mbox{set to $1$, meaning} that the next acceleration setpoint is mapped to the top end of the red line. 
    In contrast, an action of $-1$ would be mapped to the bottom end of the corresponding red line. 
    This way, each action $\in [{-1}, 1]$ leads to a feasible motion.
    Finally, a continuous trajectory is generated by linearly connecting the predicted acceleration setpoints (blue line).  
   When calculating the range of safe accelerations, it is not sufficient to focus on violations within the following interval. 
   For instance, the acceleration setpoint at $t = $ \SI{0.9}{\second} could be set to zero without causing immediate constraint violations. However, the maximum position would be exceeded at a later point in time. 
   Our approach ensures that at least one valid acceleration setpoint exists at each future decision step. 
   As a result, motions close to position or velocity limits can be learnt safely. 
   Being able to exploit the full kinematic potential of the robot joints is important for industrial automation processes as faster motions enable shorter cycle times. 
   
   In the following sections, we explain how safe accelerations are computed and demonstrate the effectiveness of our approach by learning fast motions for two different tasks without exceeding joint limits. 
   In addition, we show how the prediction frequency $f_N$ influences the learning process and transfer a policy trained in simulation to a real robot.
   Our code to compute safe accelerations is publicly available.\href{https://\linkToCode}{$^2$}

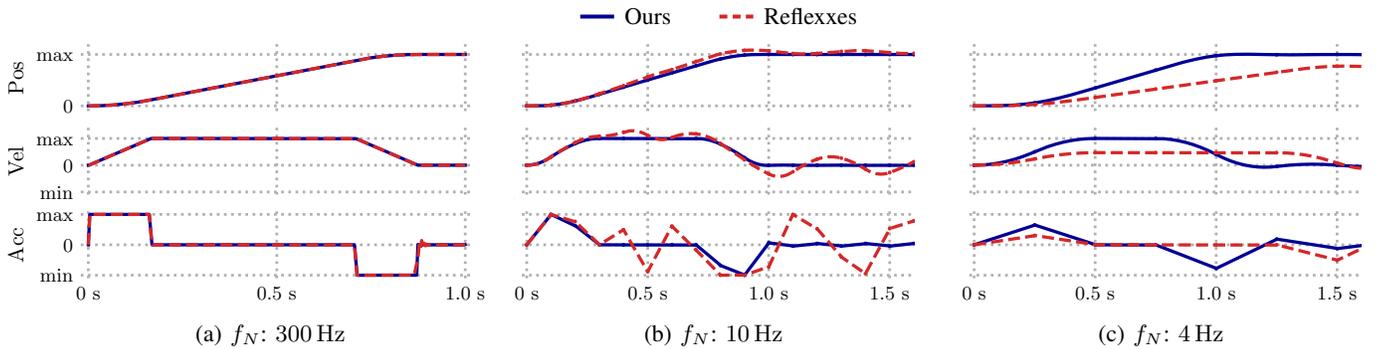
\begin{figure*}[t]
\captionsetup[subfigure]{margin=80pt}
 \begin{tikzpicture}[auto, node distance=5cm,>=latex']
        \definecolor{LINE_COLOR_B}{RGB}{214,39,40}
        \definecolor{LINE_COLOR_A}{RGB}{0,0,150}
        \node[text width=4cm] at (0, 0) (origin){};
    	\draw [draw=LINE_COLOR_A, line width=1.5pt] ($(origin.center)+(5.5cm, -0.0cm)$) -- + (0.45cm, 0cm) node[pos=1, right, yshift=-0.00cm, align=left]{\small{Ours}};
	    \draw [densely dashed, draw=LINE_COLOR_B, line width=1.5pt] ($(origin.center)+(7.35cm, 0.0cm)$) -- + (0.45cm, 0cm) node[pos=1, right, yshift=0.01cm, align=left]{\small{Reflexxes}};
    \end{tikzpicture} 
    
    \vspace{0.1cm}
    \begin{subfigure}[c]{0.33\textwidth}
	    
		\includegraphics[trim=0 0 0 5, clip, height=0.6\textwidth]{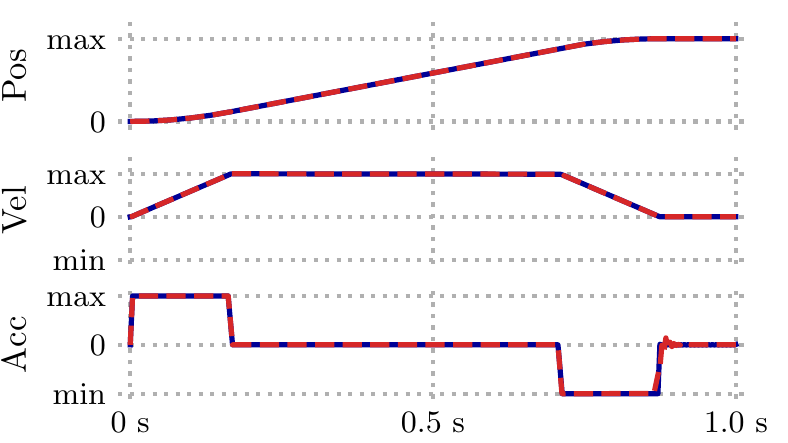}
		
	   \vspace{-0.4cm}\hspace{-1.0cm}\subcaptionbox{$f_N$: \SI{300}{\hertz}}[9cm]

	\end{subfigure}
	\hspace{0.04\textwidth}
	\begin{subfigure}[c]{0.33\textwidth}
	    \vspace{-0.0cm}
	    \includegraphics[trim=30.5 0 0 5, clip, height=0.6\textwidth]{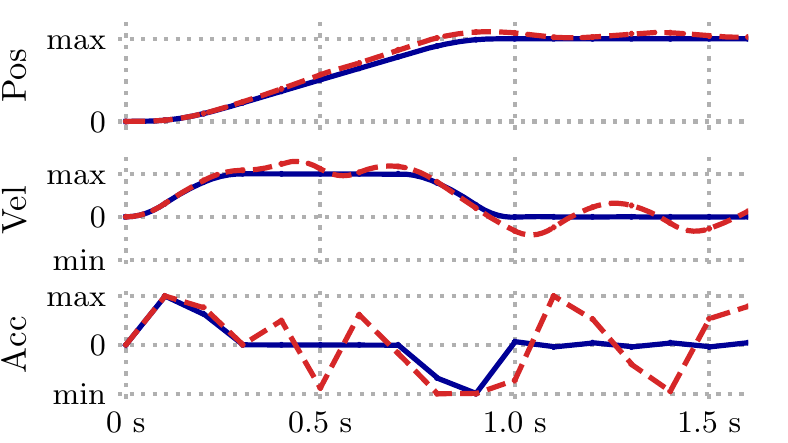}
	    
		\vspace{-0.4cm}\hspace{-1.85cm}\subcaptionbox{$f_N$: \SI{10}{\hertz}}[9cm]

	\end{subfigure} 
	\begin{subfigure}[c]{0.33\textwidth}
	    \vspace{-0.0cm}
		\includegraphics[trim=30.5 0 0 5, clip, height=0.6\textwidth]{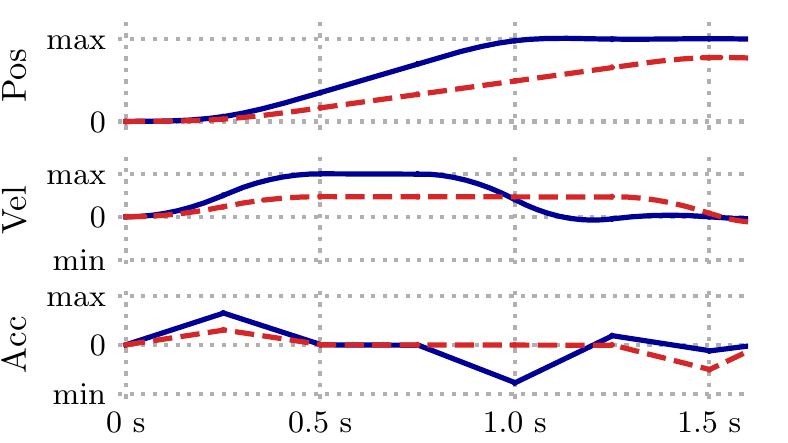}
		
	   \vspace{-0.4cm}\hspace{-1.85cm}\subcaptionbox{$f_N$: \SI{4}{\hertz}}[9cm]

	\end{subfigure}
	   
	\caption{%
	Exemplary trajectories computed with Reflexxes and with our method by selecting the largest possible acceleration at each decision step. When using a high prediction frequency (a), both approaches produce feasible trajectories. 
	If \mbox{the frequency} is reduced, the trajectory generated by Reflexxes exceeds the maximum position as in (b) or does not reach it as in (c). 
	} %
	\label{fig:reflexxes_comparison}
	\vspace{-0.25cm}
\end{figure*}
   
\section{Related work}
\label{sec:citations}
\subsection{Learning motions subject to safety constraints}
In recent years, reinforcement learning has been adapted to industrial applications like bin picking \cite{levine2018learning, kalashnikov2018scalable, berscheid2020},  autonomous assembly \cite{thomas2018learning, luo2019reinforcement} and precise insertion \cite{inoue2017deep, schoettler2019deep, kaspar2020sim2real}.
While research institutions have presented well-functioning prototypes, the usage of neural networks in industrial environments still poses major challenges in terms of safety and reliability.
According to \cite{garcia2015comprehensive}, safety constraints can be incorporated into the learning process by either modifying the optimization criterion or the exploration process. 
Modifying the optimization criterion by adding penalties \cite{andrychowicz2020learning, tan2018sim, dalal2018safe} is an easy-to-implement way to avoid undesired outcomes. However, compliance with safety constraints is not explicitly guaranteed, which is particularly problematic if the trained policy is subject to a domain transfer (e.g. training in simulation but deployment on a real robot). 

The exploration process can be influenced by incorporating external knowledge.
In some cases, the action space of a neural network can be designed in such a way that all actions are executable without violating safety constraints \cite{tan2018sim}.
If the range of valid actions depends on the current state, task-specific heuristics can be applied to adapt unsafe action selections \cite{gu2017deep}.
In \cite{dalal2018safe}, a method to learn action corrections from previous trajectories generated with random actions is presented. 
Achiam et al. \cite{achiam2017constrained} introduced a policy search algorithm for constrained Markov decision processes with near-constraint satisfaction at each iteration. In contrast, our approach provides explicit safety guarantees. %
In \cite{pham2018optlayer}, a method to consider explicit inequality constraints in the context of model-free RL is presented.  
At each decision step, a quadratic program is defined. Its solution is the closest action to an initial network prediction that satisfies the chosen constraints.
However, focusing on the current decision step is not sufficient when learning fast robot motions. As explained in the introduction, inevitable violations might occur at a later point in time. Contrary to the aforementioned work, our approach ensures that at least one valid action exists for each following decision step. 
A method to learn jerk-limited trajectories for point-to-point motions is presented in \cite{kiemel2020trueadapt}. 
However, position constraints are not considered explicitly. Instead, the trajectory execution is aborted if the deviation to a reference trajectory is higher than a specified threshold. Contrary to that, the approach presented in this paper enables safe learning of arbitrary movements without requiring precalculated reference trajectories.
\subsection{Relation to model-based trajectory optimization}
Within the context of model-based trajectory optimization and model predictive control, joint limits can be considered by defining an optimization problem with explicit inequality constraints \cite{Khatib2009Constraints}.
In contrast to model-free reinforcement learning, these techniques require a model of the dynamics and a differentiable loss function. %
While our approach considers the kinematics of robot joints, the interaction between the robot and its environment is learnt without a model. %

\subsection{Relation to online trajectory generation (OTG)}
In the context of online trajectory generation, the Reflexxes motion library \cite{kroger2011opening} can be used to compute time-optimal trajectories for industrial robots that comply with position, velocity, acceleration and jerk constraints. 
Contrary to our approach, Reflexxes assumes continuous control of the accelerations. %
Fig. \ref{fig:max_trajectory_example} shows that the maximum position is finally reached at a velocity of zero if the highest valid acceleration is selected at each decision step.   
Reflexxes expects a target position and a target velocity as input and outputs the acceleration required to reach the target state in a time-optimal way.
Given the maximum position and a velocity of zero as target, Reflexxes might be suitable to compute the highest valid acceleration.  
However, Fig. \ref{fig:reflexxes_comparison} illustrates that the library is not directly applicable to learning problems as the influence of discrete decision steps is ignored. 
At a prediction frequency of \SI{300}{\hertz}, both Reflexxes and our approach yield almost identical trajectories that do not violate the specified constraints. 
When selecting a frequency of \SI{10}{\hertz}, the trajectory generated by Reflexxes violates both position and velocity constraints. 
At a frequency of \SI{4}{\hertz}, neither the maximum position or the maximum velocity is reached when using Reflexxes. 
In conclusion, Reflexxes can be used for high prediction frequencies, while our approach ensures constraint satisfaction at arbitrary frequencies.

\begin{figure*}[t]
	\centering
    \input{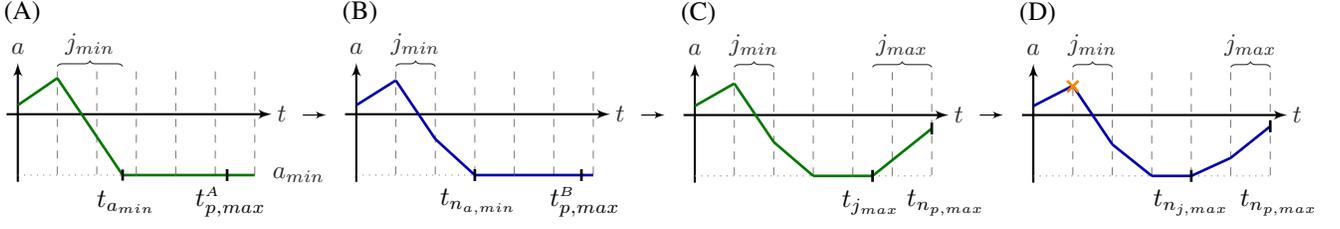}
\begin{tikzpicture}[auto, node distance=5cm,>=latex']
    	    \def\scaleFactor{0.77}
    	    \node (step_1) {\SmallPosPlot{6}{(A)}};
            \node[right of=step_1, node distance=4.5cm] (step_2) {\SmallPosPlot{5}{(B)}};
            \node[right of=step_2, node distance=4.5cm] (step_3) {\SmallPosPlot{7}{(C)}};
            \node[right of=step_3, node distance=4.5cm] (step_4) {\SmallPosPlot{8}{(D)}};

            \draw [->] ($(step_1.358)+(-0.525cm, 0cm)$) -- ($(step_2.182)+(0.025cm, 0cm)$)  node[pos=0.5, scale=\scaleFactor, above]{};
            \draw [->] ($(step_2.358)+(-0.525cm, 0cm)$) -- ($(step_3.182)+(0.025cm, 0cm)$)  node[pos=0.5, scale=\scaleFactor, above, align=center]{};
            \draw [->] ($(step_3.358)+(-0.525cm, 0cm)$) -- ($(step_4.182)+(0.025cm, 0cm)$)  node[pos=0.5, scale=\scaleFactor, above]{};
\end{tikzpicture}
	\caption{Stepwise calculation of the highest acceleration that does not lead to a violation of the position limit $p_{max}$ by successively applying the acceleration profiles (A) to (D).}
	\label{fig:profiles_steps}
\end{figure*}

\section{Problem Statement}
This work addresses the problem of learning online generation of robot motions without exceeding kinematic joint limits. 
In particular, the following constraints on the joint angle $\theta$ are defined for each robot joint and need to be satisfied at any time:
\begin{alignat}{3}
p_{min} &{}\le{}& \theta &{}\le{}& p_{max}  \label{eq:constraint_p} \\ 
v_{min} &{}\le{}& \dot{\theta} &{}\le{}& v_{max}  \label{eq:constraint_v}\\
a_{min} &{}\le{}& \ddot{\theta}&{}\le{}& a_{max}  \label{eq:constraint_a}  \\
j_{min} &{}\le{}& \dddot{\theta} &{}\le{}& j_{max},  \label{eq:constraint_j}
\end{alignat}
The limits are specified by the robot manufacturer but can also be set to lower values, e.g. to restrict the workspace of the robot or to enforce smoother movements.  
\section{Learning safe motions}
\subsection{Formalization}
We formalize the learning problem by defining a Markov decision process $(\mathcal{S}, \mathcal{A}, P_{\underline{a}}, R_{\underline{a}})$, where $\mathcal{S}$ is the state space,  $\mathcal{A}$ is the action space and $R_{\underline{a}}$ is the immediate reward caused by action $\underline{a}$. The transition probabilities $P_{\underline{a}}$ are unknown. Using model-free RL, a neural network is trained to maximize the sum of immediate rewards. Decisions are made at a prediction frequency of $f_N$. 
The immediate reward is assigned according to task-specific optimization goals. 
The state ${s_t} \in \mathcal{S}$ that is given as input to the neural network consists of two parts: 
The kinematic state of each robot joint (position $p_t$, velocity $v_t$, acceleration $a_t$) and a task-specific component $f_t$, which might include external sensor signals.
The action $\underline{\textrm{a}}_t \in \mathcal{A}$ is composed of a single scalar $m  \in [{-1}, 1]$  per joint. This scalar is translated to  $a_{t+1}$, the desired joint acceleration at the beginning of the next time interval. 
Let $a_{t+1_{S}} = [a_{t+1_{min}}, a_{t+1_{max}}]$ be the range of acceleration setpoints that does not lead to constraint violations. 
Given $a_{t+1_{S}}$, the network prediction $m$ can be mapped to a safe acceleration $a_{t+1}$:
\begin{align}
a_{t+1} = a_{t+1_{min}} + \frac{1 + m}{2} \cdot \left(a_{t+1_{max}} - a_{t+1_{min}}\right)
\end{align}
The computation is performed for each joint and the resulting movement is executed by sending linearly interpolated acceleration setpoints to a joint trajectory controller. 

\subsection{Calculation of safe accelerations}
This subsection describes, how the range of safe accelerations $a_{t+1_{S}} = [a_{t+1_{min}}, a_{t+1_{max}}]$ is computed. 
The following explanations focus on the maximum acceleration $a_{t+1_{max}}$ as the minimum acceleration $a_{t+1_{min}}$ can be calculated correspondingly. 
The basis idea is to break down the problem by calculating the maximum acceleration for each individual constraint (\ref{eq:constraint_p}) to (\ref{eq:constraint_j}). 
By selecting the smallest of these accelerations, all constraints are fulfilled simultaneously.  
In the following, the principle is explained based on position constraints.  
Further information on the other constraints and additional background knowledge can be found in \cite{kiemel2021BackgroundKnowledge}.

\newcommand{\SmallVelPlot}[2]{  
    \begin{tikzpicture}[scale=0.3]
        \def\ymax{2.5}
    	\def\ymin{-1.2}
    	\def\xdelta{2*1.6}
    	\def\xmax{3*\xdelta + 0.75}
    	\def\amin{-2.7}
    	\def\azero{0.7}
    	\def\aonemaxA{1.6}
	    \def\aonemaxB{1.45}
	    \def\jmin{-0.66*\aonemaxA} 
	    
	    \ifnum #1 < 3
	        \definecolor{POS_LIM_A}{RGB}{170,0,0}
    	\else
    	    \definecolor{POS_LIM_A}{RGB}{0,0,170}
    	\fi
    	    
    	\ifnum #1 = 1
	        \def\aonemaxA{-0.8}
    	\fi
    	
    	\ifnum #1 = 2
	        \def\aonemaxA{1.6}
    	\fi
    	
    	\ifnum #1 = 3
	        \def\aonemaxA{1.45}
    	\fi
    	    
    	\draw [<-,thick] (0,\ymax) node (yaxis) [above] {$a$} -- (0,\ymin); %
    	\draw [->,thick, name path=pathX] (-0.5, 0) -- (\xmax,0) node (xaxis) [right] {$t$};
    	
		\fill[black] (0,0) circle (1.5pt);
    	\draw[dashed] (\xdelta, \ymax)  -- (\xdelta, \ymin); %
    	\draw[dashed] (2*\xdelta, \ymax)  -- (2*\xdelta, \ymin); %
    	\draw[dashed] (3*\xdelta, \ymax)  -- (3*\xdelta, \ymin); %
    	
    	\node[color=black, text width=4cm] at ($(\xdelta,\ymax)+(2.8cm, 2.5cm)$) {#2};
    	
    	\draw[thick, color=POS_LIM_A, name path=pathOne] (0,\azero) -- (\xdelta,\aonemaxA); %

	    \ifnum #1 > 1
    	    \draw[thick, color=POS_LIM_A] (\xdelta,\aonemaxA) -- (2*\xdelta,\aonemaxA+\jmin); %
    	    \ifnum #1 = 2
    	        \draw[thick, color=POS_LIM_A, name path=pathA] (2*\xdelta,\aonemaxA+\jmin) -- (3*\xdelta,\aonemaxA+2*\jmin); %
    	    \else
        	    \draw[thick, color=POS_LIM_A, name path=pathB] (2*\xdelta,\aonemaxB+\jmin) -- (3*\xdelta, 0); %
        	\fi

        \fi
    	
    	\ifnum #1 < 3
    	    \ifnum #1 = 1
    	        \path [name intersections={of=pathOne and pathX, by=intersectionX}];
    	    \else
    	        \path [name intersections={of=pathA and pathX, by=intersectionX}];
    	    \fi
    	    \draw[thick] let \p{intersectionX}=(intersectionX) in (\x{intersectionX}, 7.5pt) -- (\x{intersectionX}, -7.5pt) node[pos=1, below=0.13] {$t_{a_0}'$};
    	\fi
    	\ifnum #1 = 3
    	    \draw[thick] (2*\xdelta, 7.5pt) -- (2*\xdelta, -7.5pt) node[pos=1, below=0.13] {$t_{n_{a,0}}$};
    	\fi

    \end{tikzpicture} 
}

Fig. \ref{fig:profiles_steps} illustrates the steps to calculate the highest possible acceleration at the next decision step that does not lead to a violation of the position limit $p_{max}$. %
The desired acceleration is marked by an orange cross.
A robot joint exceeds its position limit if the maximum position $p_{max}$ is reached at a velocity greater than zero. 
On the contrary, the highest possible acceleration  can be achieved if the maximum position is reached at a velocity of zero. 
Assuming $t_0 = 0$, this condition can be expressed as follows: 
\begin{align}
v_0 + \int_{0}^{t_{p,max}}  \! a(t) \ dt &= 0  \label{eq:int_bounded_pos_1} \\[1ex]
p_0 + v_0 \cdot t_{p,max} + \int\displaylimits_{0}^{t_{p,max}} \! \! \int\displaylimits_{0}^{t} a(t) \ dt \, dt  &= p_{max}, \label{eq:int_bounded_pos_2} 
\end{align}
with $t_{p,max}$ being the time at which the maximum position is reached. 
Equations (\ref{eq:int_bounded_pos_1}) and (\ref{eq:int_bounded_pos_2}) can be solved for the desired acceleration if the course of future accelerations is known. %
Given that the desired acceleration should be as high as possible, it has to be followed by a deceleration phase. The deceleration should be as strong as possible, but has to comply with the jerk and acceleration limits. %
As visualized in Fig. \ref{fig:profiles_steps}, the required acceleration profile can be determined in a step-wise manner: 
\begin{itemize}
\item In step (A), the minimum jerk $j_{min}$ is applied until the minimum acceleration $a_{min}$ is reached. By solving equations (\ref{eq:int_bounded_pos_1}) and (\ref{eq:int_bounded_pos_2}), the continuous switching time $t_{a_{min}}$ can be calculated. Since the decisions of our neural network are made at discrete time steps, this profile is not directly applicable to our problem. However, $t_{n_{a,min}}$, the time of the decision step that follows $t_{a_{min}}$, can be computed.   
\item Knowing $t_{n_{a,min}}$, the acceleration profile shown in step (B) can be applied. This profile is used to calculate $t_{p,max}^{\scriptscriptstyle B}$, the time at which the maximum position is reached. %
As demonstrated in the accompanying video\footnote[3]{\url{https://youtu.be/JpkKCd9jyss}}, the use of this profile induces oscillations that can be avoided if the maximum position is reached at a discrete decision step. Given $t_{p,max}^{\scriptscriptstyle B}$, the next discrete time step $t_{n_{p,max}}$ can be computed. 
\item If $t_{n_{p,max}}$ is known, oscillations can be suppressed by applying acceleration profile (C). %
Similar to step (A), a continuous switching time $t_{j_{max}}$ is computed in the first place. Based on this value, the discrete switching time $t_{n_{j,max}}$ is calculated. 
\item Given $t_{n_{j,max}}$, the desired maximum acceleration, which is marked by an orange cross, can finally be calculated by applying acceleration profile (D).  
\end{itemize}

\subsection{Implementation}

For given acceleration profiles, like those shown in Fig. \ref{fig:profiles_steps}, specific systems of equations can be derived from (\ref{eq:int_bounded_pos_1}) and (\ref{eq:int_bounded_pos_2}).
For each resulting equation system, an analytical solution is obtained using SymPy\cite{meurer2017sympy}. %
To speed up calculations, the analytical expressions are translated into C code and compiled as a Python module. %
Our step-wise approach supports arbitrary prediction frequencies and does not rely on numeric solvers. In addition, our analytical expressions are differentiable. We note that the range of safe accelerations $a_{t+1_{S}}$ can also be translated into a range of safe velocities $v_{t+1_{S}}$ or positions $p_{t+1_{S}}$. 
Mapping the predicted actions linearly along $a_{t+1_{S}}$, $v_{t+1_{S}}$ or $p_{t+1_{S}}$ is equivalent.

\section{Evaluation}
\label{sec:result}

\subsection{Description of the evaluation tasks}
We evaluate our approach by applying it to two model-free learning problems performed by a KUKA iiwa robot with seven degrees of freedom. 
Both tasks are simulated with PyBullet \cite{coumans2016pybullet} and trained using an RL algorithm called Proximal Policy Optimization \cite{schulman2017proximal}.
Renderings  of the resulting policies can be found in the accompanying video\footnotemark[3]. 
\subsubsection{Velocity maximization task}
\begin{figure}[t]
    \begin{subfigure}[c]{0.23\textwidth}
	    \vspace{0.1cm}
		\includegraphics[trim=0 22 0 10, clip, width=\textwidth]{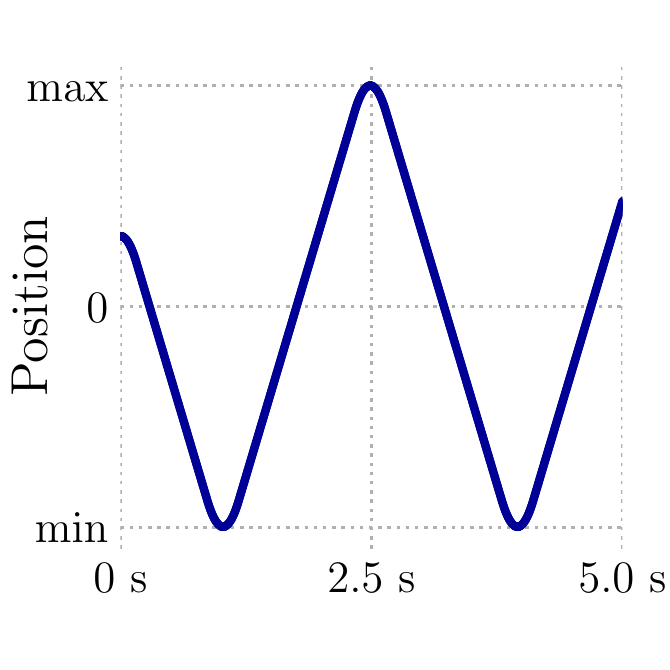}

	   \subcaption{Optimal trajectory for the velocity maximization task.}
	   \label{fig:evaluation_tasks_a}
		
	\end{subfigure}
	\hspace{0.01\textwidth}
	\begin{subfigure}[c]{0.22\textwidth}
	    \vspace{-0.0cm}
	    \includegraphics[trim=350 230 300 170, clip, width=\textwidth]{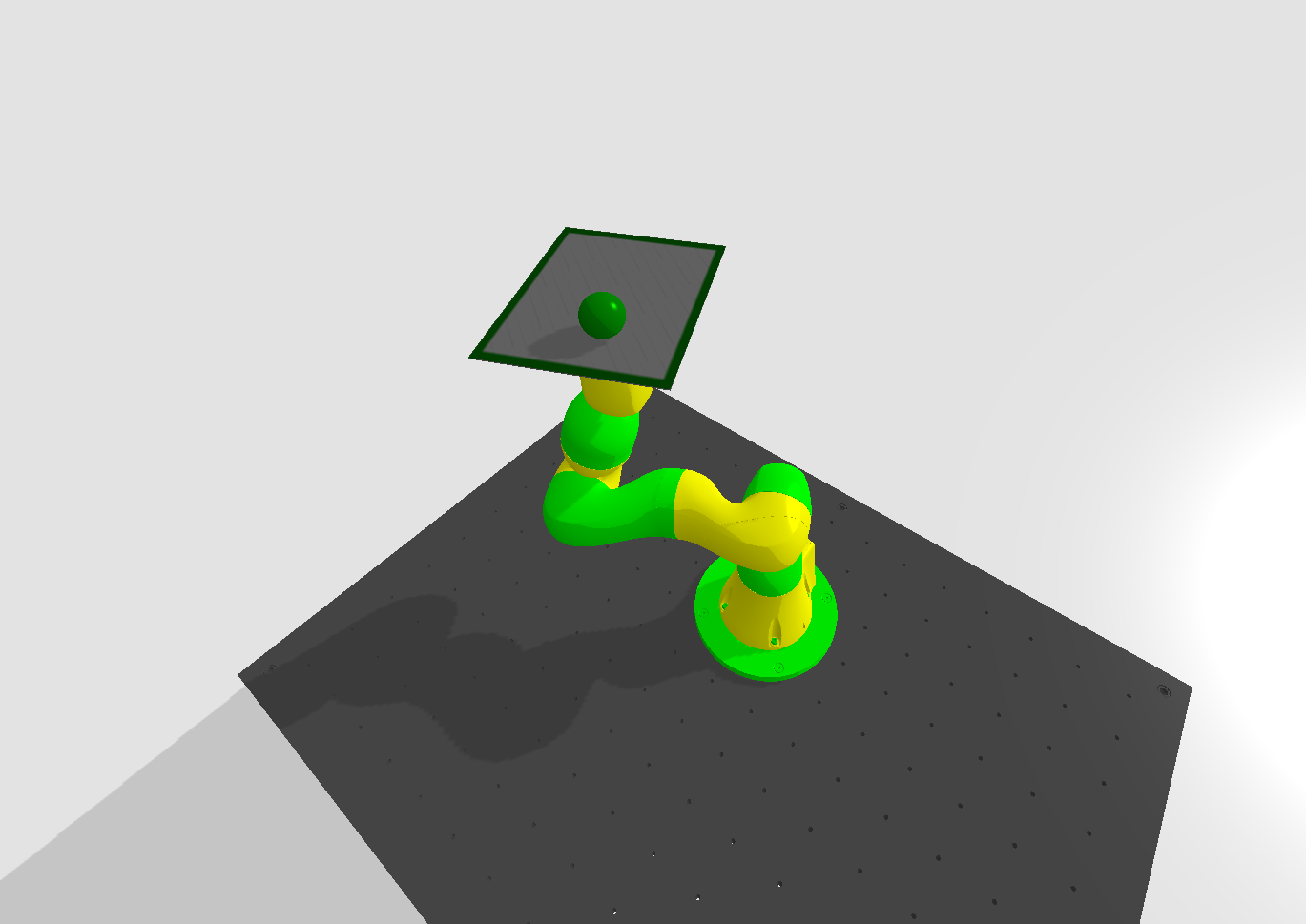}
	    
		\vspace{0.1cm}
		\subcaption{Ball-on-plate task}
		\label{fig:evaluation_tasks_b}

	\end{subfigure} 
	   
	\caption{Evaluation tasks} %
	\label{fig:evaluation_tasks}
	\vspace{-0.35cm}
\end{figure}

The goal of this task is to maximize the absolute velocity of all joints over time without violating kinematic constraints.
As shown in Fig. \ref{fig:evaluation_tasks_a}, the optimal performance is achieved if each joint oscillates between its upper and its lower position limit. This task is selected for our evaluation, as the robot joints are required to work close to their kinematic limits in order to receive high rewards. In addition, the average absolute velocity achieved by the optimal policy helps to assess the final performance of the learning process. %
The duration of each generated trajectory is set to five seconds. 

\subsubsection{Ball-on-plate task}
In this task, the robot is trained to roughly follow a reference trajectory while balancing a ball on a plate.
The following two varieties differ with respect to their reward function:
The goal of the first version is to keep the ball as close as possible to its initial position (``in place"), while the second variant aims to prevent the ball from falling off the plate (``on plate"). Further information on the ball-on-plate task can be found in \cite{kiemel2020trueadapt}.

\subsection{Constraint satisfaction}
\begin{figure}[b]
	\includegraphics[trim=0 15 0 0, clip, width=\linewidth]{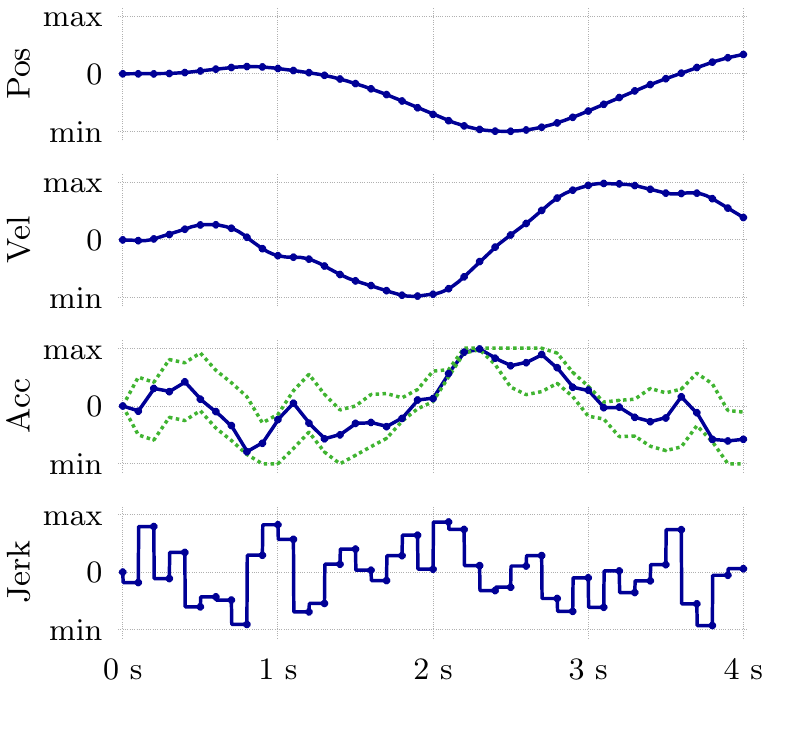}
	\caption{
	 An exemplary trajectory generated by choosing random actions. Our method ensures that joint constraints are never violated.  
	 Prediction frequency  $f_N$: \SI{10}{\hertz}}
	\label{fig:max_random_trajectories}
\end{figure}

As a first step, we evaluate whether all trajectories generated by our method satisfy the specified constraints \mbox{(\ref{eq:constraint_p}) to (\ref{eq:constraint_j})}.
Fig. \ref{fig:max_random_trajectories} shows an exemplary trajectory for a single joint that is produced by choosing random \mbox{actions $\in [{-1}, 1]$} at each decision step. %
The green dotted lines represent the range of safe accelerations. %
As expected, the kinematic limits are not violated. 
At $t\!=\!\SI{2.2}{\second}$, the range of safe accelerations is very small. However, at least one valid acceleration setpoint exists at any decision step. 
The border case of selecting the largest possible acceleration at each decision step is shown in Fig. \ref{fig:max_trajectory_example} on the front page of this paper.
It can be seen that the resulting trajectory is constrained by the maximum jerk, the maximum acceleration, the minimum jerk, the maximum velocity and finally by the maximum position.
To assess the reliability of our approach on a larger scale, two neural networks are trained for the velocity maximization task.
The prediction frequencies of the networks are set to \SI{240}{\hertz} and \SI{20}{\hertz}, respectively. 
We generate 1000 trajectories with each network and determine the highest occurring position and velocity value. 
These values are normalized with respect to their limits and listed in Table \ref{table:reward_shaping}. 
For both frequencies, the maximum position and the maximum velocity are equal to the specified limits.

\begin{table}[t]
  \caption{Performance metrics for the velocity maximization task using different strategies to avoid constraint violations. Each settings is evaluated by generating 1000 episodes with neural networks that are trained until convergence.}
    \makegapedcells
\begin{tabular*}{0.49\textwidth}{p{2mm}p{20mm}p{10mm}p{12mm}p{10mm}p{10mm}} 
    \toprule
& Method & Average \newline velocity & Violation \newline  rate & Max. \newline  position & Max. \newline  velocity \\
    \hline
    
& Ours
    & $\textcolor{TABLE_GOOD}{0.91}$ & \SI[color=TABLE_GOOD]{0.0}{\percent}    & $\textcolor{TABLE_GOOD}{1.00}$   & $\textcolor{TABLE_GOOD}{1.00}$      \\
    
& Soft penalties
     & $\textcolor{TABLE_BAD}{0.51}$ & \SI{8.0}{\percent}    & $1.25$   & $0.90$    \\
     
\rot{\rlap{\SI{240}{\hertz}}} 
&  Hard penalties
     &  \SI{0.71} & \SI[color=TABLE_BAD]{52.7}{\percent}    & $\textcolor{TABLE_BAD}{1.43}$   & $\textcolor{TABLE_BAD}{1.56}$  \\

     \hline
& Ours
    & $\textcolor{TABLE_GOOD}{0.88}$ & \SI[color=TABLE_GOOD]{0.0}{\percent}    & $\textcolor{TABLE_GOOD}{1.00}$   & $\textcolor{TABLE_GOOD}{1.00}$     \\
    
& Soft penalties
     & $\textcolor{TABLE_BAD}{0.44}$ & \SI{1.2}{\percent}    & $0.93$   & $1.14$    \\
     
\rot{\rlap{\SI{20}{\hertz}}}
&  Hard penalties
     & \SI{0.58} & \SI[color=TABLE_BAD]{41.5}{\percent}    & $\textcolor{TABLE_BAD}{1.14}$   & $\textcolor{TABLE_BAD}{1.69}$     \\

    \bottomrule
    \end{tabular*}
    \vspace{1ex}
\label{table:reward_shaping}
\end{table}
\begin{table}[b]
 \caption{Performance metrics for the velocity maximization task at different prediction frequencies.}
    \makegapedcells
\begin{tabular*}{0.49\textwidth}{@{}p{11mm}p{8mm}p{8mm}p{9mm}p{12mm}p{12mm}} 
    \toprule
\multirow[t]{2}{*}{Frequency}
    & \multicolumn{3}{l}{Average velocity}
         & \multicolumn{2}{l}{Trajectories until convergence} \\ %
     & Random & Trained & Optimal
            & Absolute & Relative                                     \\
    \hline
\SI{240}{\hertz}
    & $\textcolor{TABLE_BAD}{0.212}$ & $\textcolor{TABLE_GOOD}{0.912}$    & $\textcolor{TABLE_GOOD}{0.921}$   & $\textcolor{TABLE_GOOD}{48\,900}$ & $\textcolor{TABLE_GOOD}{1.00}$     \\
    
\SI{120}{\hertz}
     & $0.268$    & $0.908$    & $0.920$ & $62\,650$ & $1.28$    \\
     
\SI{20}{\hertz}
     & $0.431$    & $0.880$    & $0.919$   & $107\,250$ & $2.20$    \\

\SI{10}{\hertz}
     & $\textcolor{TABLE_GOOD}{0.487}$    & $\textcolor{TABLE_BAD}{0.865}$    & $\textcolor{TABLE_BAD}{0.913}$   & $\textcolor{TABLE_BAD}{128\,250}$ & $\textcolor{TABLE_BAD}{2.62}$    \\
    \bottomrule
    \end{tabular*}
\label{table:evalution_frequency_velocity_maximization}
\end{table}
\subsection{Benchmarking with reward shaping}
Avoiding constraint violations by adding penalties to the reward function is an easy-to-implement alternative to our approach. 
Table \ref{table:reward_shaping} compares our method with two different versions of penalty assignment based on the learning performance of the velocity maximization task. %
Green and red values indicate the best and the worst outcome for each metric.  
When using hard penalties, the reward is set to zero if a position or a velocity constraint is violated. 
In case of soft penalties, the reward is gradually reduced if either the current position or the current velocity is higher than \SI{50}{\percent} of its maximum value.
The violation rate indicates the fraction of trajectories with at least one constraint violation.   
Our results show that both penalty-based techniques fail to prevent constraint violations. 
While the violation rate with soft penalties is comparatively low, penalty-based methods do not provide explicit safety guarantees, which is unacceptable when working with real robots. 
In addition, the average velocity, which should be maximized as part of the learning process, is significantly higher when using our method.
The results are plausible as penalties distort the desired optimization target.
In this particular case, the penalties encourage the robot to stay away from the joint limits, which is in contradiction to the goal of the velocity maximization task.

\subsection{Influence of the prediction frequency $f_N$}
Our approach supports arbitrary prediction frequencies as the impact of discrete decision steps is considered.
In this subsection, the influence of the prediction frequency on the learning process is analyzed. 
Table \ref{table:evalution_frequency_velocity_maximization} lists various performance metrics for the velocity maximization task. %
The results show that both the optimal and the trained policy achieve higher velocities if the prediction frequency is increased. 
This behaviour is plausible as higher prediction frequencies allow more granular control of the robot joints.  
\begin{table}[t]
    \caption{Performance metrics for the ball-on-plate task. An episode is considered as successful if the reference trajectory is tracked and if the ball is balanced correctly. } %
    \makegapedcells
\begin{tabular*}{0.49\textwidth}{p{2mm}p{22mm}p{16mm}p{16mm}p{16mm}} 
    \toprule

   & Frequency & Success rate & Balancing \newline error & Tracking \newline error    \\
    \hline
& No adaptations
    &  \SI{0.6}{\percent} & \SI[color=TABLE_BAD]{99.4}{\percent}    & \SI[color=TABLE_GOOD]{0.0}{\percent} \\
    
& \SI{240}{\hertz}
    & \SI{0.5}{\percent} & \SI{11.0}{\percent}    & \SI[color=TABLE_BAD]{88.5}{\percent}  \\
    
& \SI{120}{\hertz}
     & \SI{81.1}{\percent} & \SI{3.5}{\percent}    & \SI{15.4}{\percent} \\
     
& \SI{20}{\hertz}
     & \SI[color=TABLE_GOOD]{99.4}{\percent} & \SI[color=TABLE_GOOD]{0.1}{\percent}    & \SI{0.5}{\percent} \\

\rot{\rlap{In place}} & \SI{2.5}{\hertz}
     & \SI[color=TABLE_BAD]{0.0}{\percent} & \SI{88.2}{\percent}    & \SI{11.8}{\percent}   \\
     \hline
     
& No adaptations
    & \SI[color=TABLE_BAD]{4.7}{\percent} & \SI[color=TABLE_BAD]{95.3}{\percent}    & \SI[color=TABLE_GOOD]{0.0}{\percent}   \\

& \SI{240}{\hertz}
    & \SI{33.5}{\percent} & \SI{64.2}{\percent}    & \SI[color=TABLE_BAD]{2.3}{\percent} \\
    
& \SI{120}{\hertz}
     & \SI{85.8}{\percent} & \SI{14.1}{\percent}    & \SI{0.1}{\percent}  \\
     
& \SI{20}{\hertz}
     & \SI[color=TABLE_GOOD]{98.5}{\percent} & \SI[color=TABLE_GOOD]{1.4}{\percent}    & \SI{0.1}{\percent}  \\

\rot{\rlap{On plate}} & \SI{2.5}{\hertz}
     & \SI{94.8}{\percent} & \SI{4.8}{\percent}    & \SI{0.4}{\percent} \\     
     
    \bottomrule
    \end{tabular*}
    \vspace{1ex}
\label{table:evalution_frequency_ball_on_plate}
\end{table}
\begin{figure}[b]
\begin{tikzpicture}[auto, node distance=5cm,>=latex']
        \definecolor{LINE_COLOR_B}{RGB}{214,39,40}
        \definecolor{LINE_COLOR_A}{RGB}{0,0,150}
        \node[text width=4cm] at (0, 0) (origin){};
    	\draw [draw=LINE_COLOR_A, line width=1.5pt] ($(origin.center)+(0.3cm, 0.0cm)$) -- + (0.3cm, 0cm) node[pos=1, right, yshift=-0.00cm, align=left]{\small{Setpoints}};
	    \draw [densely dashed, draw=LINE_COLOR_B, line width=1.5pt] ($(origin.center)+(2.6cm, 0.0cm)$) -- + (0.3cm, 0cm) node[pos=1, right, yshift=0.01cm, align=left]{\small{Actual values}};
    \end{tikzpicture}
    \vspace{-0.0cm}
	\begin{subfigure}[c]{0.23\textwidth}
	    \vspace{-0.0cm}
		\includegraphics[trim=0 0 0 10, clip, height=0.9\textwidth]{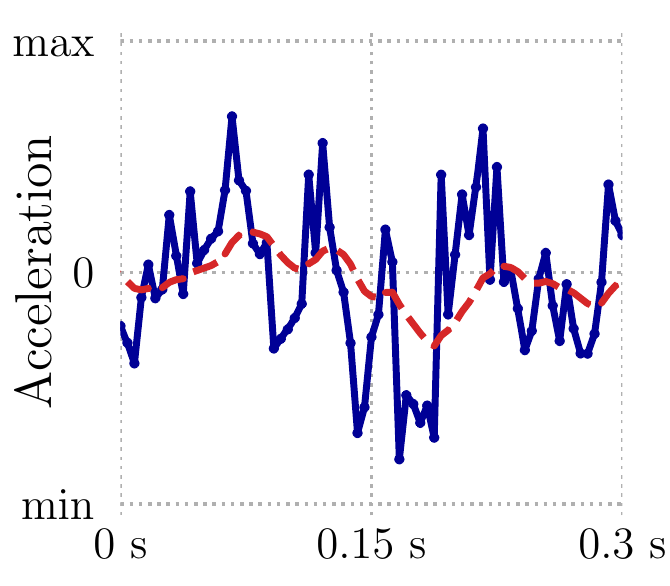}
	    	
	   \vspace{-0.6cm}\hspace{0.9cm}\subcaptionbox{$f_N$: \SI{240}{\hertz}}[3cm]

		\label{fig:vel_limitation_a}
		
	\end{subfigure}
	\hspace{0.016622\textwidth}
	\begin{subfigure}[c]{0.23\textwidth}
	    \vspace{-0.0cm}
	    \includegraphics[trim=27.5 0 0 10, clip, height=0.9\textwidth]{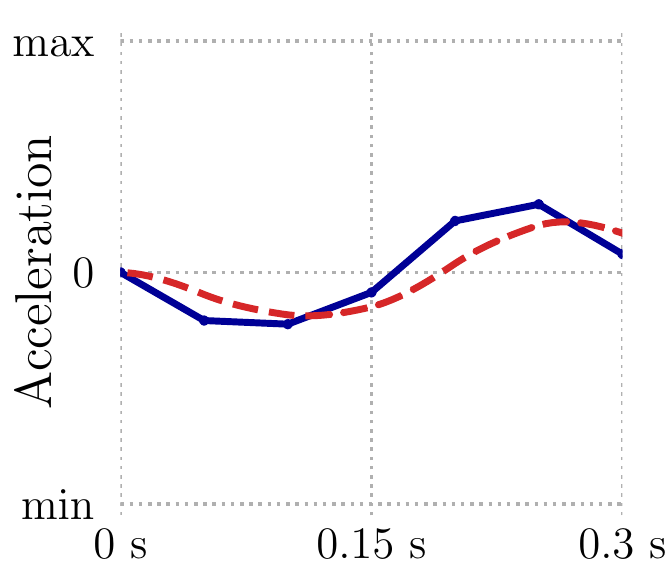}
	    
		 \vspace{-0.6cm}\hspace{0.2cm}\subcaptionbox{$f_N$: \SI{20}{\hertz}}[3cm]

		\label{fig:vel_limitation_b}
		
	\end{subfigure}
	\caption{Exemplary acceleration setpoints compared to actual values for a single joint.}
	\label{fig:acc_setpoint_actual}
\end{figure}
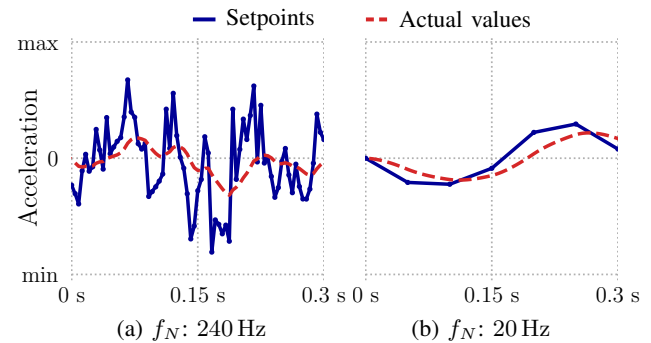
To compare the data efficiency, we assume convergence of a training process once the gain in reward has reached \SI{98}{\percent} of its final value. 
Our results show that the number of trajectories required for convergence decreases if the prediction frequency is increased. 
Data efficiency is especially important if the training process is to be performed by a real robot.
Our observations are plausible as higher prediction frequencies lead to more training data per trajectory.
Table \ref{table:evalution_frequency_ball_on_plate} shows the performance of the ball-on-plate task at different prediction frequencies. 
The best balancing performance is achieved at a prediction frequency of \SI{20}{\hertz}. 
If the prediction frequency is set to \SI{2.5}{\hertz}, the ``in place" version fails as the time between decision steps is too high to prevent the ball from moving.
Contrary to the velocity maximization task, the ball-on-plate task is influenced by the tracking performance of the trajectory controller.
Fig. \ref{fig:acc_setpoint_actual} shows exemplary acceleration setpoints and the resulting actual values for a trajectory segment of the ball-on-plate task.  %
Since the trajectory controller behaves like a low-pass, the acceleration setpoints are tracked more accurately, if the prediction frequency is reduced. %
Beyond that, the reaction of the system is delayed due to the inertia of the ball. 
At high prediction frequencies, the correlation between the predicted action and the reaction of the system is harder to assess. 
We conclude that the final learning performance does not necessarily profit from higher prediction frequencies if the controlled system behaves similarly to a low-pass.%

\subsection{Computing effort and real-time capability}
\begin{figure}[t]
    \begin{tikzpicture}[scale=1.0]
            \def\scaleFactor{0.88}
             \def\ymax{0.75}
        	\def\ymin{0}
        	\def\azero{0.6}
        	\def\xdelta{7.6}
        	\def\xmax{1*\xdelta + 0.5}
        
        	\def\sensorDataPos{0.27}
        	\def\sensorDataTextPos{0.41}
        	\def\actionPredictionPos{0.55}
        	\def\actionPredictionTextPos{0.66}
        	\def\safetyChecksPos{0.77}
        	\def\safetyChecksTextPos{0.865}
        	\def\ticklength{5pt}

        	\definecolor{POS_LIM_A}{RGB}{170,0,0} %
        	\definecolor{POS_LIM_B}{RGB}{0,0,150}
        	\definecolor{POS_LIM_C}{RGB}{0,90,0}
        	\definecolor{POS_LIM_D}{RGB}{197,43,11}
        	\definecolor{POS_LIM_E}{RGB}{35,0,74}
    
    		\draw [->,thick, name path=pathX] (-0.25, 0) -- (\xmax,0) node[pos=\sensorDataPos] (sensorData) {}  node[pos=\sensorDataTextPos] (sensorDataText) {} node[pos=\actionPredictionPos] (actionPrediction) {} node[pos=\actionPredictionTextPos] (actionPredictionText) {} node[pos=\safetyChecksPos] (safetyChecks) {} node[pos=\safetyChecksTextPos] (safetyChecksText) {}  node (xaxis) [right] {$t$};
    		
        	\draw [-{Latex[scale=1.0]},thick, color=black] (0*\xdelta,\ymin) node[below=0.05, black] {$t_0$} -- (0*\xdelta,\ymax); %
        	\draw [-{Latex[scale=1.0]},thick, color=black] (1*\xdelta,\ymin) node[below=0.05, black] {$t_1$} -- (1*\xdelta,\ymax); %
        	
        	\draw[thick] ($(sensorData.center)+(0, \ticklength)$) -- ($(sensorData.center)-(0, \ticklength)$);
        	\draw[thick] ($(actionPrediction.center)+(0, \ticklength)$) -- ($(actionPrediction.center)-(0, \ticklength)$);
        	\draw[thick] ($(safetyChecks.center)+(0, \ticklength)$) -- ($(safetyChecks.center)-(0, \ticklength)$);
	        
	        \node[color=POS_LIM_C, align=center,  scale=\scaleFactor] at ($(sensorDataText.center)+(0,0.43cm)$) {Sensor data \\collection};
	        \node[color=POS_LIM_B, align=center,  scale=\scaleFactor] at ($(actionPredictionText.center)+(0,0.40cm)$) {Action \\prediction};
        	\node[color=POS_LIM_A, align=center,  scale=\scaleFactor] at ($(safetyChecksText.center)+(-0.1cm, 0.40cm)$) {Safety \\checks};
        	
        	\def\braceXOffset{0.00}
        	\def\braceYOffset{0.1}
        	
        	\draw[black,decorate, decoration={brace,amplitude=4pt}] ($(sensorData.center)+(0,\ymax + \braceYOffset)$) -- (\xdelta, \ymax + \braceYOffset)  node[midway, above,yshift=4pt, xshift=0.0cm]{$\Delta t_J$};

    	\end{tikzpicture}
    \caption{Jitter $\Delta t_J$ caused by the processing time of real systems.}
    \label{fig:jitter}
\end{figure}
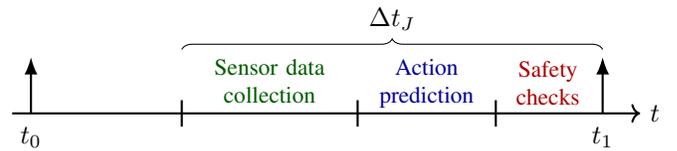

In order to transfer a control policy from simulation to a real robot, all calculations need to be real-time capable. 
This subsection analyzes the calculating times to generate trajectories measured on a system equipped with a CPU (Intel i7-8700K) and a GPU (Nvidia GTX 1080 Ti). 
Computing the range of safe accelerations $a_{t+1_{S}}$ took \SI{111}{\micro\second} per joint and decision step, which is comparable to the calculating time required by Reflexxes (\SI{123}{\micro\second}). %
Table \ref{table:computation_times} shows how the computing effort is influenced by the selected prediction frequency.  
\begin{table}[b]
\caption{Calculating times to generate a trajectory with a duration of 1000 seconds.}

\makegapedcells
\begin{tabular*}{0.49\textwidth}{@{}p{9mm}p{5.0mm}p{5mm}p{2mm}p{5mm}p{5mm}p{6mm}p{5mm}} 
    \toprule
\multirow[t]{2}{*}{Frequency}
& \multirow[t]{2}{*}{Total}
 & \multicolumn{2}{l}{Action prediction}
    & \multicolumn{2}{l}{Safety checks}
    &  \multicolumn{2}{l}{Other calculations} 
     \\
    & & Abs. & Rel.  & Abs. & Rel.  & Abs. & Rel. \\
    \hline
\SI{240}{\hertz}
    & \SI[color=TABLE_BAD]{544}{\second} & \SI[color=TABLE_BAD]{194}{\second}    & \SI{36}{\percent}    & \SI[color=TABLE_BAD]{187}{\second}    & \SI{34}{\percent} & \SI[color=TABLE_BAD]{163}{\second}    & \SI{30}{\percent}   \\
    
\SI{120}{\hertz}
     & \SI{284}{\second} & \SI{96}{\second}    & \SI{34}{\percent}    & \SI{93}{\second}    & \SI{33}{\percent} & \SI{95}{\second}    & \SI{33}{\percent}   \\
     
\SI{20}{\hertz}
    & \SI{71}{\second} & \SI{16}{\second}    & \SI{23}{\percent}    & \SI{16}{\second}    & \SI{23}{\percent} & \SI{39}{\second}    & \SI{54}{\percent}   \\

\SI{10}{\hertz}
     & \SI[color=TABLE_GOOD]{49}{\second} & \SI[color=TABLE_GOOD]{8}{\second}    & \SI{16}{\percent}    & \SI[color=TABLE_GOOD]{8}{\second}    & \SI{16}{\percent} & \SI[color=TABLE_GOOD]{33}{\second}    & \SI{68}{\percent}   \\
    \bottomrule
    \end{tabular*}
    \vspace{1ex}
\label{table:computation_times}
\end{table}
The computing effort increases at higher prediction frequencies as the calculations have to be performed more frequently. 
It is important to note that the calculating times need to be significantly shorter than the execution time of the resulting trajectory.
Ideally, all calculations should be made at discrete points in time as illustrated by the arrows at $t_0$ and $t_1$ in Fig. \ref{fig:jitter}. In practice, sensor data has to be collected prior to the desired decision time, which leads to a timing jitter of $\Delta t_J$. The relative jitter $J_R = \frac{\Delta t_J}{t_1 - t_0}$ increases at higher prediction frequencies and is further raised if additional safety checks for collision avoidance are to be performed. 
In summary, the technical effort required to achieve real-time capability is reduced  when choosing a lower prediction frequency. 
As shown in the accompanying video, we successfully transferred a policy for the ball-on-plate task from simulation to a real robot using a prediction frequency of \SI{20}{\hertz}. Since our network predicts accelerations rather than torques, the domain transfer could be conducted without further measures to compensate model errors. %

\section{Conclusion and future work}
\label{sec:conclusion}
This paper presented an approach to learn fast robot trajectories while satisfying kinematic joint constraints. The effectiveness of our method is demonstrated by successfully learning two different tasks without violating joint limits.
In contrast to penalizing constraint violations, our approach provides explicit safety guarantees, which is crucial when working with real robots. 
The proposed method considers the impact of discrete decision steps, thereby enabling arbitrary prediction frequencies.
Our experiments showed that the prediction frequency influences the final learning performance, the amount of training data required until convergence and the processing power needed for real-time execution.  %

For practical applications, it is also important to avoid collisions and violations of torque limits when generating robot movements. 
Therefore, the coupling between the robot joints has to be taken into account. In our follow-up work \cite{kiemel2021torque}, we present a method for learning torque-limited robot trajectories while avoiding collisions with static obstacles and other robots. In future work, we are interested in extending our approach such that collisions with moving obstacles can also be avoided.

\section*{ACKNOWLEDGMENT}

This research was supported by the German Federal Ministry of Education and Research (BMBF) and the Indo-German Science \& Technology Centre (IGSTC) as part of the project TransLearn (01DQ19007A). We thank Tamim Asfour for his valuable feedback and advice.

\bibliographystyle{IEEEtran}
\bibliography{root}

\end{document}